\documentclass[10pt]{article}
\usepackage{PRIMEarxiv}
\usepackage{adjustbox}
\usepackage{longtable}
\usepackage{url}
\usepackage{needspace}
\usepackage[numbers, sort&compress]{natbib}
\usepackage{graphicx, xcolor, todonotes, appendix, booktabs, caption, array, authblk, svg}
\usepackage{listings}

\title{MDCrow: Automating Molecular Dynamics Workflows with Large Language Models}

\author[1]{Quintina Campbell$^\dagger$}
\author[1,3]{Sam Cox$^\dagger$}
\author[1]{Jorge Medina$^\dagger$}
\author[2]{Brittany Watterson}
\author[1,3]{Andrew D. White$^*$}

\affil[1]{Department of Chemical Engineering, University of Rochester, Rochester, New York, USA}
\affil[2]{Department of Biomedical Engineering,
University of Rochester, Rochester, New York, USA}
\affil[3]{FutureHouse Inc., San Francisco, CA}

\begin{document}
\date{\today}

\maketitle
\def\thefootnote{$\dagger$}\footnotetext{These authors contributed equally to this work}
\def\thefootnote{*}\footnotetext{Corresponding author: \texttt{andrew.white@rochester.edu}}
\def\thefootnote{\arabic{footnote}}

\begin{abstract}
Molecular dynamics (MD) simulations are essential for understanding biomolecular systems but remain challenging to automate. Recent advances in large language models (LLM) have demonstrated success in automating complex scientific tasks using LLM-based agents. In this paper, we introduce MDCrow, an agentic LLM assistant capable of automating MD workflows. MDCrow uses chain-of-thought over 40 expert-designed tools for handling and processing files, setting up simulations, analyzing the simulation outputs, and retrieving relevant information from literature and databases. We assess MDCrow's performance across 25 tasks of varying required subtasks and difficulty, and we evaluate the agent's robustness to both difficulty and prompt style. \texttt{gpt-4o} is able to complete complex tasks with low variance, followed closely by \texttt{llama3-405b}, a compelling open-source model. While prompt style does not influence the best models' performance, it has significant effects on smaller models.
\end{abstract}
\section{Introduction} 

Molecular dynamics (MD) simulations is a common method to understand dynamic and complex systems in chemistry and biology. While MD is now routine, its integration into and impact on scientific workflows has increased dramatically over the past few decades \cite{MD_Applications_in_proteins, MD_review_2002, hollingsworth2018molecular}. There are two main reasons for this: First, MD provides valuable insights. Through simulations, scientists can study structural and dynamic phenomena, perturbations, and dynamic processes in their chemical systems. Second, innovations in hardware and expert-designed software packages have made MD much more accessible to both experienced and novice users \cite{hollingsworth2018molecular}. 

For a given protein simulation, parameter selection is nontrivial: the user must provide the input structure (such as a PDB \cite{velankar2021protein} file), select a force field (e.g., CHARMM \cite{brooks2009charmm}, AMBER \cite{ponder2003force}), and specify parameters such as temperature, integrator, simulation length, and equilibration protocols. Simulations also generally require pre- and post-processing steps, along with various analyses. For instance, a user may need to clean or trim a PDB file, add a solvent, or analyze the protein's structure. After simulation, they might examine the protein's shape throughout the simulation or assess its stability under different conditions. The choices for pre-processing, analysis, and simulation parameters are highly specific to any given use case and often require expert intuition. Thus, automating this process is difficult but beneficial. 

Several efforts have been made to automate MD workflows \cite{Admiral, RadonPy,Gmx_qk,Foundry,Chaperong,MolAr,ProFESSA,Sim_stack,Fab_sim,PyAuto_FEP,Proto_Caller}, focusing largely on specific domains, such as RadonPy for polymer's simulations \cite{RadonPy}, or PyAutoFEP for proteins and small molecules for drug-screening \cite{PyAuto_FEP}. Other approaches are constrained to a particular combination of simulation software and simulation (e.g. GROMACS and Free Energy Perturbation). Certainly, there has been significant community-driven improvement in automating and creating MD toolkits \cite{suplatov2020easyamber, martinez2009packmol, michaud2011mdanalysis, mdtraj, eastman2017openmm, abraham2015gromacs, lammps,Sim_stack} and user-friendly interfaces and visualizations \cite{goret2017mdanse, ribeiro2018qwikmd, rusu2014mdwiz, hildebrand2019bringing, biarnes2012metagui, humphrey1996vmd, sellis2009gromita, martinez2017playmolecule}.
While these advances improve the capabilities and ease of use in many cases, the inherent variability of MD workflows still poses a great challenge for full automation. 

Large-Language Model (LLM) agents \cite{schick2023toolformer, mrkl, react, narayanan2024aviary} have gained popularity for their ability to automate technical tasks through reasoning and tool usage, even surpassing domain-specialized LLMs (e.g., BioGPT \cite{luo2022biogpt}, Med-PaLM \cite{singhal2023large}) when programmed for specialized roles \cite{gao2024empoweringbiomedicaldiscovery}. These agents have demonstrated promising results in scientific tasks within a predefined toolspace, with tools like ChemCrow and Coscientist successfully automating complex workflows and novel design in chemical synthesis \cite{bran2024augmenting, boiko2023autonomous, cactus}. Likewise, LLM-driven automation has been explored in materials research \cite{jablonka202314, su2024automation, chiang2024llamp, kim2024large}, literature and data aggregation \cite{lee2024harnessing, skarlinski2024language}, and more sophisticated tasks \cite{calms, BioPlanner, crisprgpt, LLM-RDF, chiang2024llamp, ChatMOF, TAIS, ProtAgent}. Most similar to this work, ProtAgents \cite{ProtAgent} is a multi-agent modeling framework tackling protein-related design and analysis, and LLaMP \cite{chiang2024llamp} applies a retrieval-augmented generation (RAG)-based ReAct agent to simulate inorganic materials by interfacing with literature databases, Wikipedia, and atomistic simulation tools. Although preliminary work has applied agentic LLMs to MD via a RAG-based agent \cite{chiang2024llamp}, no fully adaptive and autonomous system exists for biochemical MD or protein simulations. See Ramos et al.\cite{ramos2024review} for a recent review on the design, assessment, and applications of scientific agents.

\begin{figure}[ht!]
    \centering
    \includegraphics[width=0.9\linewidth]{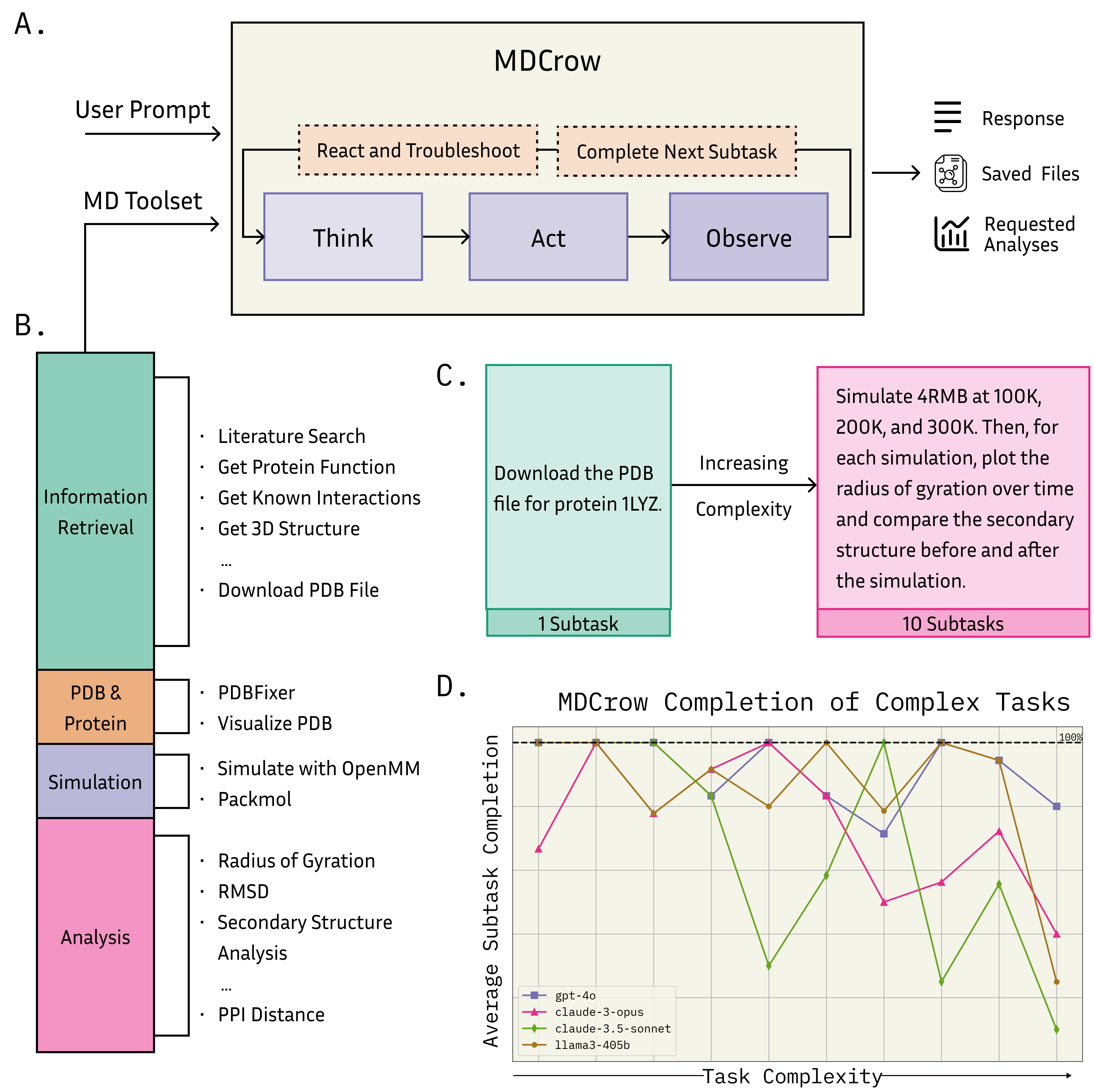}
    \caption{\textbf{A.} MDCrow workflow. Starting with a user prompt and initialized with a set of MD tools, MDCrow follows a chain-of-thought process until it completes all tasks in the prompt. The final output includes a response, along with all resulting analyses and files. \textbf{B}. The tool distribution categorized into 4 types: information retrieval, PDB and protein handling, simulation, and analysis. A few examples from each category are shown. \textbf{C.} Two example prompts that MDCrow is tested on. The first is the simplest prompt, containing only 1 subtask. The most complex task requires 10 subtasks. \textbf{D}. Average subtask completion across all 25 prompts as task complexity (the number of subtasks per prompt) increases. The top three performing base-LLMs are shown. Among them, \texttt{gpt-4o} and \texttt{llama3-405b} consistently maintain high stability, staying close to 100\% completion even as task complexity increases.
    }
    \label{fig:toc}
\end{figure}

Here we present MDCrow, an LLM-agent capable of autonomously completing MD workflows. Our main contributions to the field are (1) we assess MDCrow's performance across 25 tasks with varying difficulty and compare performance of different LLM models; (2) we measure robustness how agents are prompted and task complexity based on required number of subtasks we compare with simply equipping an LLM with a python interpreter with the required packages installed, rather than using a custom built environment. Our main conclusions is that MDCrow with \texttt{gpt-4o} or \texttt{llama3-405b} is able to perform nearly all of our assessed tasks and is relatively insensitive to how precise the instructions are given to it. See Figure \textbf{\ref{fig:toc}D} for an overview of the main results.

\section{Methods} \label{methods}

\subsection{MDCrow Toolset}
MDCrow is an LLM agent, which consists of an environment of tools that emit observations and an LLM that selects actions (tools + inpnut arguments). MDCrow is built with Langchain \cite{langchain} and a ReAct style prompt.\cite{react}. The tools mostly consist of analysis and simulation methods; we use OpenMM \cite{eastman2017openmm} and MDTraj \cite{mdtraj} packages, but in principle our findings generalize to any such packages. 

MDCrow's tools can be categorized in four groups: Information Retrieval, PDB \& Protein, Simulation, and Analysis (see Figure \textbf{\ref{fig:toc}B}). 

\paragraph{Information Retrieval Tools} These tools enable MDCrow to build context and answer simple questions posed by the user. Most of the tools serve as wrappers for UniProt API functionalities \cite{uniprot}, allowing access to data such as 3D structures, binding sites, and kinetic properties of proteins. Additionally, we include a LiteratureSearch tool, which uses PaperQA \cite{skarlinski2024language} to answer questions and retrieve information from literature. PaperQA accesses a local database of relevant PDFs, selected specifically for the test prompts, which can be found in SI section \textbf{\ref{references_table}}. This real-time information helps the system provide direct answers to user questions and can also assist the agent in selecting parameters or guiding simulation processes. 

\paragraph{PDB \& Protein Tools} MDCrow uses these tools to interact directly with PDB files, performing tasks such as cleaning structures with PDBFixer \cite{eastman2017openmm}, retrieving PDBs for small molecules and proteins, and visualizing PDBs through Molrender \cite{molrender} or NGLview \cite{nguyen2018nglview}.

\paragraph{Simulation Tools} All included simulation tools use OpenMM \cite{eastman2017openmm} for simulation and PackMol \cite{martinez2009packmol} for solvent addition. These tools are built to manage dynamic simulation parameters, handle errors related to inadequate parameters or incomplete preprocessing, and address missing forcefield templates efficiently. The agent responds to simulation setup errors through informative error messages, improving overall robustness. Finally, the simulation tools outputs Python scripts that can be modified directly by MDCrow whenever the simulation requires additional steps or parameters.

\paragraph{Analysis Tools} This group of tools is the largest in the toolset, designed to cover common MD workflow analysis methods, many of which are built on MDTraj \citep{mdtraj} functionalities. Examples include computing the root mean squared distance (RMSD) with respect to a reference structure, the radius of gyration, analyzing the secondary structure, and various plotting functions.

\subsection{Chatting with Simulations}
\begin{figure}[ht!]
    \centering
    \includegraphics[width=0.8\linewidth]{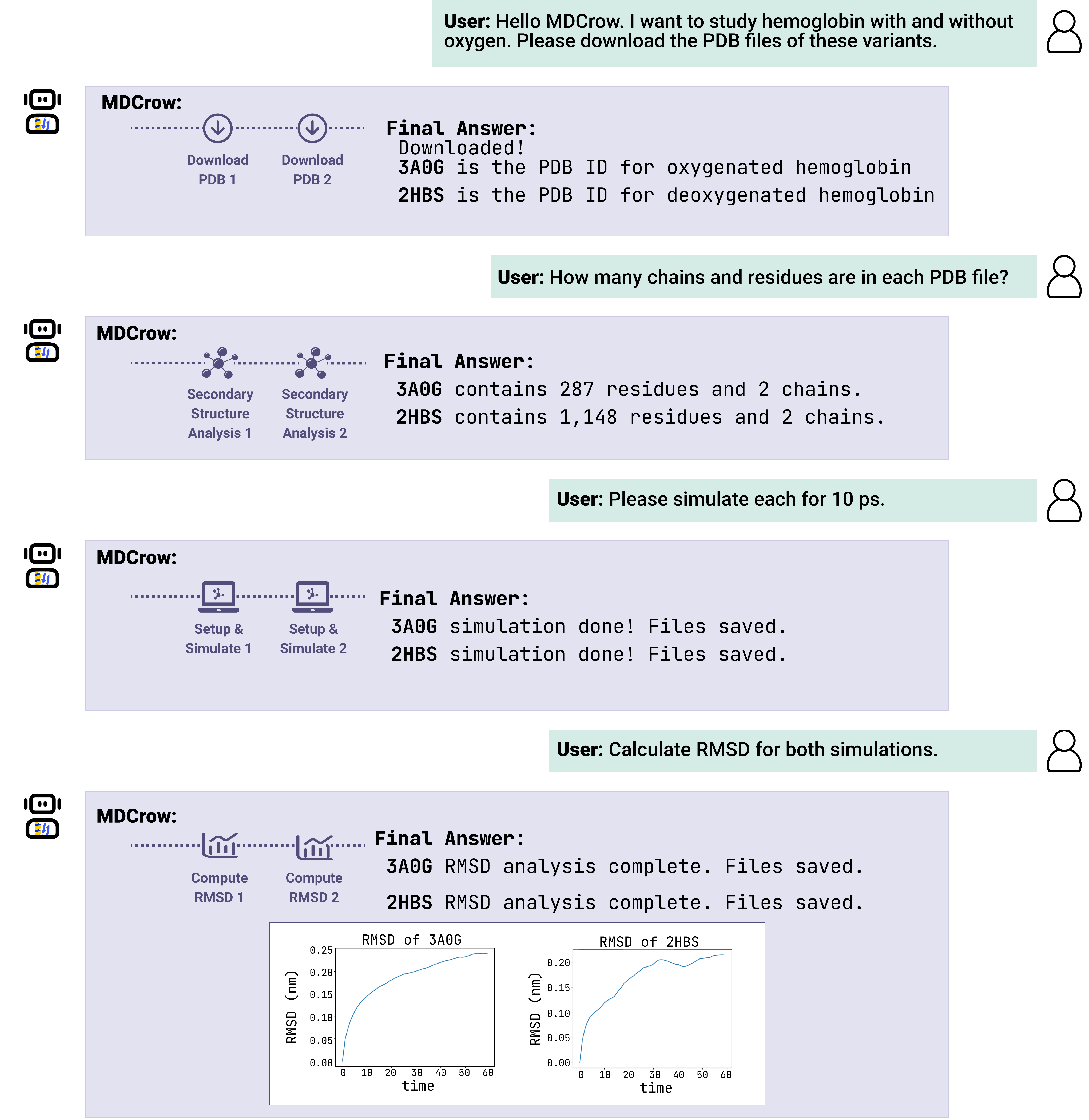}
    \caption{\textbf{Example Chat} Example of chat with MDCrow. The user first asks to download PDB files for two systems. Then, once MDCrow has completed this task, the user asks for analysis of the files. Next, the user asks for a quick 10 ps simulation of both files, and MDCrow saves all files for later handling. Lastly, the user asks for plots of RMSD for each simulation over time, and MDCrow responds with each plot.}
    \label{fig:chat}
\end{figure}

A key challenge in developing an automated MD assistant is ensuring it can manage a large number of files, analyses, and long simulations and runtimes. Although MDCrow has been primarily tested with shorter simulations, it is designed to handle larger workflows as well. Its ability to retrieve and resume previous runs allows users to start a simulation, step away during the long process, and later continue interactions and analyses without needing to stay engaged the entire time. An example of this chatting feature is shown in Figure \textbf{\ref{fig:chat}}.

MDCrow creates an LLM-generated summary of the user prompt and agent trace, which is assigned to a unique run identifier provided at the end of the run (but accessible at any time during the session). Each run's files, figures, and path registry are saved in a unique checkpoint folder linked to the run identifier.

When resuming a chat, the LLM loads the summarized context of previous steps and maintains access to the same file corpus, as long as the created files remain intact. To resume a run, the user simply provides the checkpoint directory and run identifier. MDCrow then loads the corresponding memory summaries and retrieves all associated files, enabling seamless continuation of analyses.

\section{Results}
\subsection{MDCrow Performance on Various Tasks}
To assess MDCrow's ability to complete tasks of varying difficulty, we designed 25 prompts with different levels of complexity and documented the number of subtasks (minimum required steps) needed to complete each task. MDCrow was not penalized for taking additional steps, but was penalized for omitting necessary ones. For example, the first prompt in Figure \textbf{\ref{fig:toc}C} contains a single subtask, whereas the complex task requires 10 subtasks: downloading the PDB file, performing three simulations, and performing two analyses per simulation. If the agent failed to complete an earlier step, it was penalized for every subsequent step it could not perform due to that failure.

\begin{figure}[ht!]
    \centering
    \includegraphics[width=1.0\linewidth]{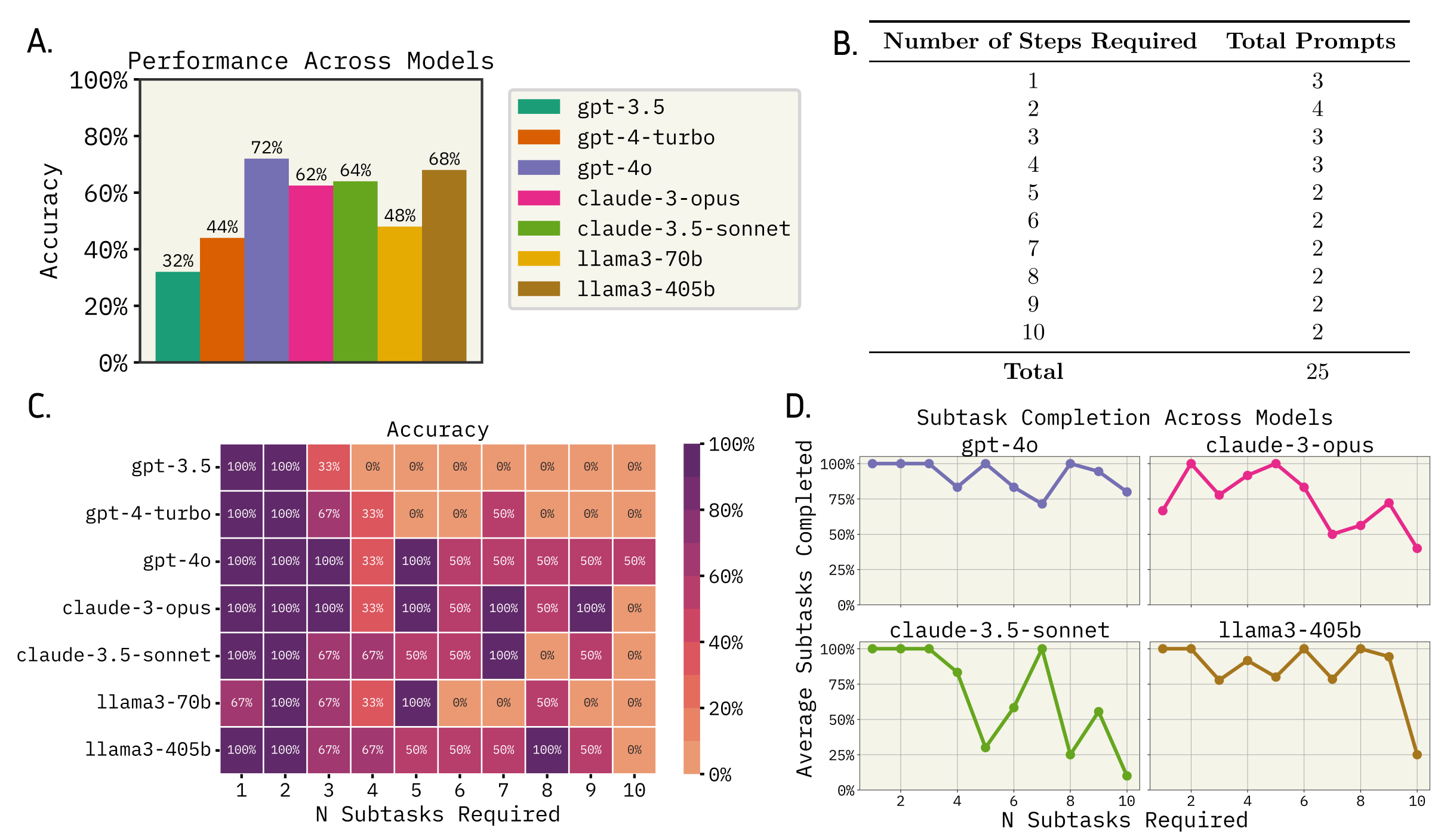}
    \caption{MDCrow Performance across Large Language Models.
    \textbf{A.} Summary of MDCrow performance dependent on LLM. Percentage of accuracy is determined by whether it gave acceptable final answer or not. While statistically indistinguishable from Claude and Llama models, \texttt{gpt-4o} significantly outperforms the rest of GPT models on giving accurate solutions (t-test, $0.004 \le$ p-value $\le 0.046$).  % p values for gpt4o beating gpt-3.5 and gpt-4-turbo: 0.00396 and 0.0459 respectively
    % NOT significant -- p values for gpt4o beating opus 3, sonnett 3.5, llama3-70b, llama3-405b: 0.96, 0.69, 0.08, 0.76, respectively
    \textbf{B.} The distribution of number of subtasks in each task across 25 prompts. The prompts range from 1-10 steps, with each step count belonging to at least 2 prompts. 
    \textbf{C.} Percentages of prompts with accurate solutions with respect to LLM used and number of subtasks per task. The correlation between accuracy and complexity is statistically significant for all LLMs (Spearman correlation,  $3.9\times10^{-7} \le$ p-value $\le 1.1\times10^{-2}$)
    % gpt-3.5 (0.0000049), gpt-4-turbo (0.00000039), gpt-4o (0.00011), Claude 3 Opus (0.011), Claude 3.5 Sonnet (0.000051), llama3-70b (0.000074), llama3-405b (0.00027)
    \textbf{D.} Percentage of the subtasks that the agent completed for each base LLM per task. 
    }
    \label{fig:results_1}
\end{figure}

The 25 prompts require between 1 and 10 subtasks, with their distribution shown in Figure \textbf{\ref{fig:results_1}B}. Each prompt was tested across three GPT models (\texttt{gpt-3.5-turbo-0125, gpt-4-turbo-2024-04-09, gpt-4o-2024-08-06}) \cite{gpt3, achiam2023gpt}, two Llama models (\texttt{llama-v3p1-405b-instruct, llama-v3p1-70b-instruct}) \cite{dubey2024llama} (accessed via the Fireworks AI API with 8-bit floating point (8FP) quantization \cite{fireworksAI2024}), and two Claude models (\texttt{claude-3-opus-20240229, claude-3-5-sonnet-20240620}) \cite{sonnet_model_card, opus_model_card}. A newer Claude Sonnet model, \texttt{claude-3-5-sonnet-20241022} was tested in later experiments but was not found to give superior results, so it was not tested on these 25 prompts. All other parameters were held constant across tests, and each version of MDCrow executed a single run per prompt.

Each run was evaluated by experts recording the number of required subtasks the agent completed and using Boolean indicators to indicate accuracy, whether the agent triggered a runtime error, and whether the trajectory contained any hallucinations. Since the agent trajectories for each run are inherently variable, accuracy is defined as the result's consistency with the expected trajectory rather than comparing against a fixed reference.

The percentage of tasks that were deemed to have valid solutions for MDCrow with each base-LLM is shown in Figure \textbf{\ref{fig:results_1}A}. The lowest performing model was \texttt{gpt-3.5}. This is not surprising, as this model had some of the highest hallucination rates (32\% of prompt completions contained hallucinations), compared to the absence of documented hallucinations in the higher performing models, \texttt{gpt-4o} and \texttt{llama3-405b}. However, the discrepancy in accuracy rates between models cannot solely be attributed to hallucinations, as \texttt{gpt-3.5} attempted fewer than half of the required subtasks, whereas the higher-performing models, \texttt{gpt-4o} and \texttt{llama3-405b}, attempted 80-90\% of the required subtask, earning accuracy in answering for 72\% and 68\% of tasks, respectively (Figures \textbf{\ref{fig:results_1}C, D}).

These results indicate that MDCrow can handle complex MD tasks but is limited by the capabilities of the base model. For \texttt{gpt-4-turbo}, \texttt{gpt-3.5}, and \texttt{llama3-70b}, the number of trajectories with verified results decreases significantly as task complexity increases (Figure \textbf{\ref{fig:results_1}C}). In contrast, \texttt{gpt-4o} and \texttt{llama3-405b} show only a slight decline, demonstrating that MDCrow performs well even for complex tasks when paired with more robust base models.

\subsection{MDCrow Robustness}
\begin{figure}
    \centering
    \includegraphics[width=1.0\linewidth]{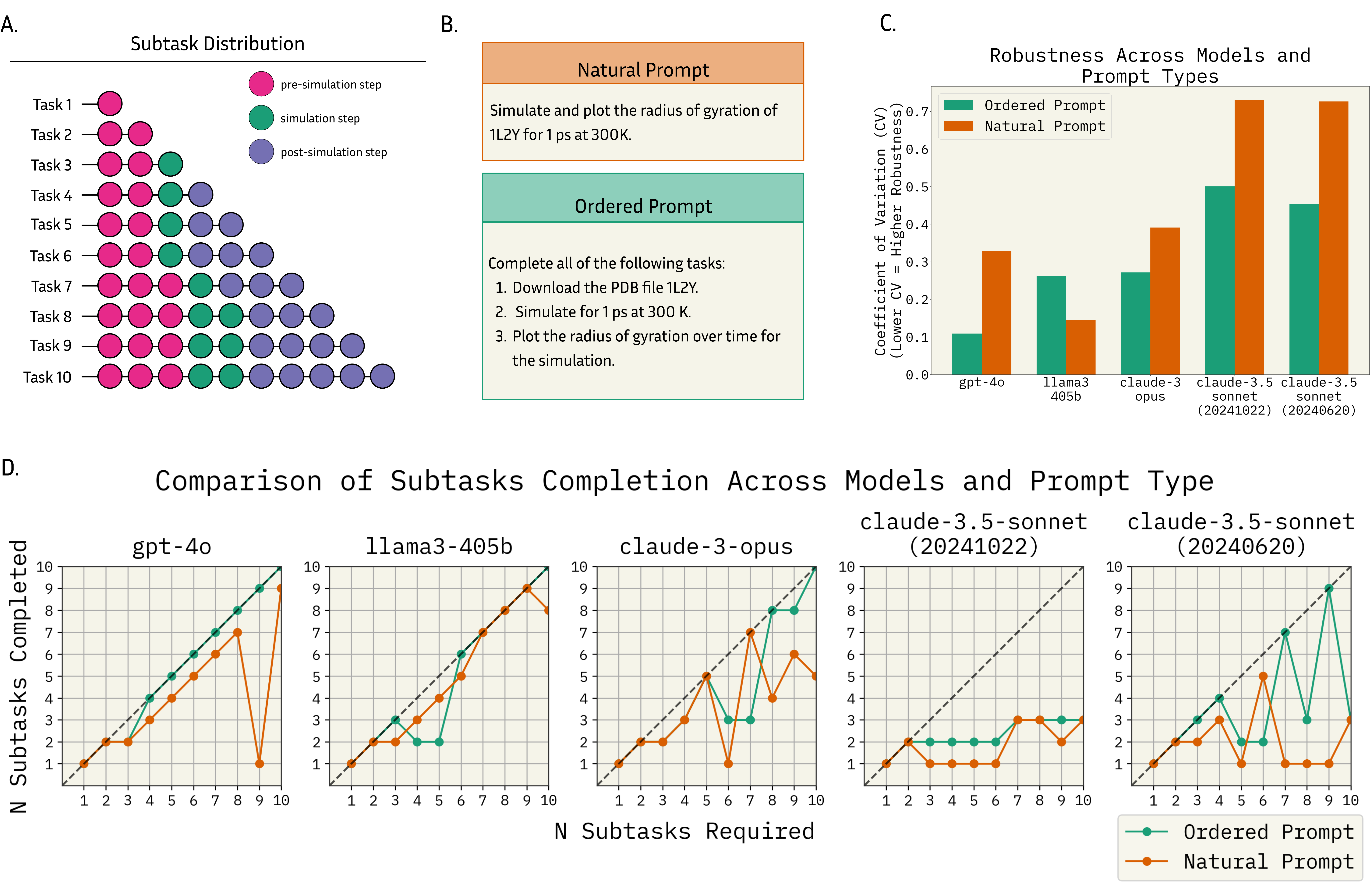}
    \caption{\textbf{A}. The number of subtasks in each task, categorized by type. Task 1 begins with a single pre-simulation subtask (\textit{Download a PDB file}) and each subsequent task adds a single subtask, adding to a total of 10 tasks with a maximum of 10 subtasks. \textbf{B}. Example of "Natural" and "Ordered" prompt style on a three-step prompt. \textbf{C}. The robustness of MDCrow built on each model with both prompt types, measured by coefficient of variation (CV). Lower CV is interpreted as greater consistency. \texttt{gpt-4o} and \texttt{llama3-405b} are the more robust models, as the Claude models have higher CVs. \textbf{D}. Comparison of subtask completion across models and prompt types. In the 9-subtask prompt, \texttt{gpt-4o} encountered an error after only one step and gave up without trying to fix it. In general, \texttt{gpt-4o} and {llama3-405b} have relatively robust performance with increasing complexity for both prompt types. \texttt{claude-3-opus} struggles with more complex tasks, making more logical errors for complex tasks. The two \texttt{claude-3.5-sonnet} models showed fairly poor performance across this experiment.}
    \label{fig:results_robustness}
\end{figure}

We evaluated the robustness of MDCrow on complex prompts and different prompt styles. We hypothesized that some models would excel at completing complex tasks, while others would struggle—either forgetting steps or hallucinating—as the number of required subtasks increased. To test this, we created a sequence of 10 prompts that increased in complexity. The first prompt required a single subtask, and each subsequent prompt added an additional subtask (see Figure \textbf{\ref{fig:results_robustness}A}). Each prompt was tested twice: once in a natural, conversational style and once with explicitly ordered steps. Example prompts can be seen in Figure \textbf{\ref{fig:results_robustness}B}.

To quantify robustness, we calculated the coefficient of variation (CV) for the percentage of completed subtasks across tasks. A lower CV indicates greater consistency in task completion and, therefore, higher robustness. The analysis revealed clear differences in robustness across models and prompt types. Overall, \texttt{gpt-4o} and \texttt{llama3-405b} demonstrated moderate to high robustness, while the Claude models showed significantly lower robustness. The performance comparison is shown in Figure \textbf{\ref{fig:results_robustness}C}.

We expected that the percentage of subtasks completed by each model would decrease as task complexity increased. However, with \texttt{gpt-4o} and \texttt{llama3-405b} as base models, MDCrow demonstrated a strong relationship between the number of required and completed subtasks (Figure \textbf{\ref{fig:results_robustness}D}) for both prompt types, indicating consistent performance regardless of task complexity or prompt style. The three included Claude models demonstrated less impressive performance. \texttt{claude-3-opus} followed the linear trend very loosely, becoming more erratic as task complexity increased. As the tasks required more subtasks, the model consistently misses nuances in the instructions and make logical errors. Both \texttt{claude-3.5-sonnet} models gave poor performance on these tasks, often producing the same error (see SI section \textbf{\ref{Claude-SI}}).

\subsection{MDCrow Comparison}
We also compared MDCrow to two baselines: a ReAct \cite{react} agent with only a Python REPL tool and a single-query LLM. MDCrow and the baselines were tested on the same 25 prompts as previously mentioned, all using \texttt{gpt-4o}. We use different system prompts to accommodate each framework, guiding the LLM to utilize common packages with MDCrow, and these prompts can be found in SI section \textbf{\ref{prompts-SI}}. 

\begin{figure}[ht!] 
    \centering
    \includegraphics[width=1.0\linewidth]{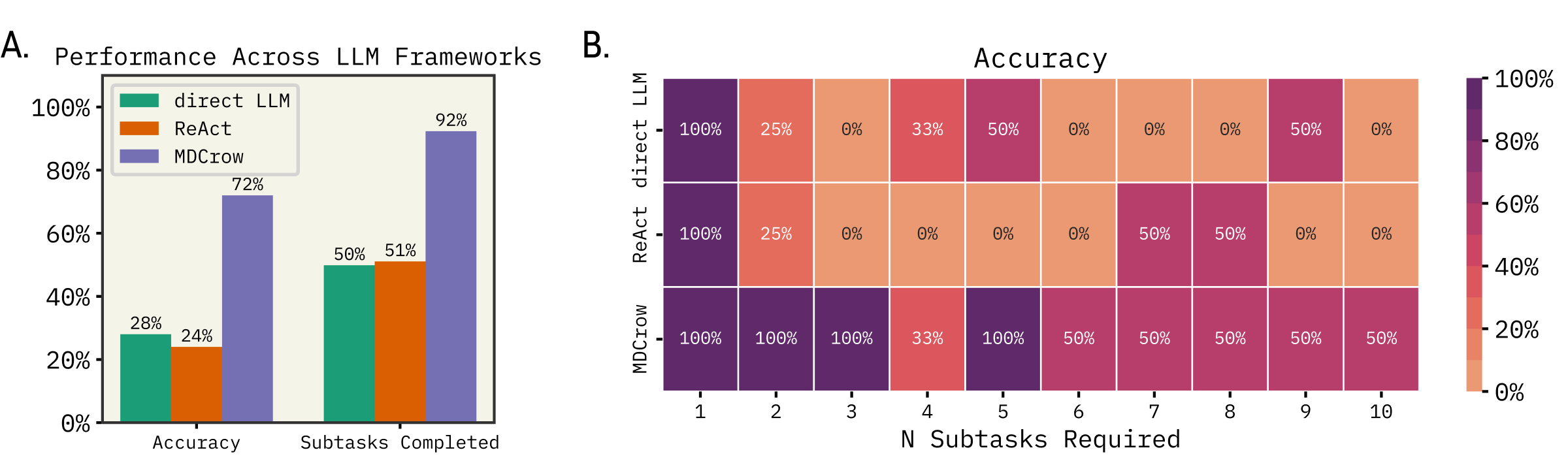}
    \caption{Performance across LLM Frameworks using the same 25-prompt set: MDCrow, direct LLM with no tools (single-query), and ReAct agent with only Python REPL tool. All use \texttt{gpt-4o}. 
    \textbf{A.} Performance among LLM frameworks measured by whether accuracy and average percentage of subtasks they complete for each of 25 task prompts. 
    MDCrow is significantly better at giving accurate solutions than direct LLM (t-test, $p=1\times10^{-3}$) and ReAct (t-test, $p=4\times10^{-4}$).
    MDCrow completes significantly more subtasks on average compared to direct LLM (t-test, $p=1\times10^{-6}$) and ReAct (t-test, $p=6\times10^{-6}$).
    \textbf{B.} Percentage of tasks completed with the respect to LLM framework used and the number of subtasks required for each task. The correlation between accuracy and number of subtasks required is statistically significant, %$p<0.05$ 
    $p=1\times10^{-3}$ for direct LLM and $p=1\times10^{-4}$ MDCrow. The p value for ReAct is $p=7\times10^{-2}$. %$p<0.10$
    }
    \label{fig:llm_fw_exps}
\end{figure}

The single-query LLM is asked to complete the prompt by writing the code for all subtasks, not unlike what standalone ChatGPT would be asked to do. We then execute the code ourselves and evaluate the outcomes accordingly. ReAct with Python REPL can write and execute codes using a chain-of-thought framework.
We find that MDCrow outperforms the two baselines significantly, as shown in Figure \textbf{\ref{fig:llm_fw_exps}A}, on attempting all subtasks and achieving an accurate solution. Not surprisingly, the two baseline methods struggled with code syntax errors and incorrect handling of PDB files. There is not a significant difference between the two baselines, indicating that the ReAct framework did not significantly boost the model's robustness. 

In Figure \textbf{\ref{fig:llm_fw_exps}B}, we observe that the performance of all three methods generally declines as task complexity increases. However, both baseline methods drop to zero after just three steps, with performance then fluctuating erratically at higher complexities. This is not surprising, as proper file processing and simulation setup are crucial for optimal LLM performance in MD tasks. In contrast, MDCrow demonstrates greater robustness and reliability in handling complex tasks, thanks to its well-designed system for accurate file processing and simulation setup, as well as its ability to dynamically adjust to errors.

\subsection{MDCrow Extrapolation through Chatting}
\begin{figure}[h!]
    \centering
    \includegraphics[width=\linewidth]{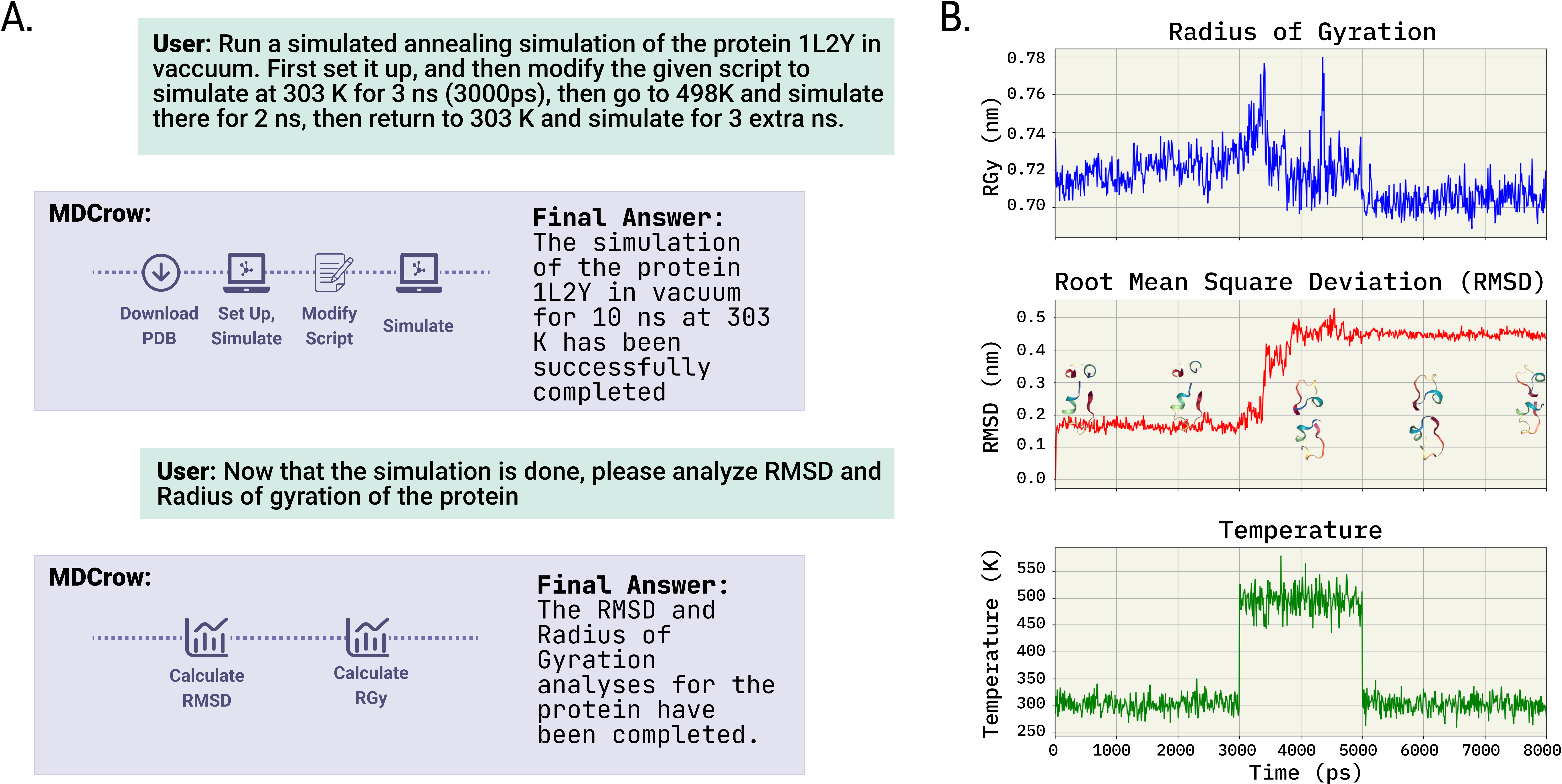}
    \caption{
    \textbf{A.} MDCrow simulating annealing. The user directly instructs MDCrow to simulate an annealing simulation of protein 1L2Y. Once the simulation is complete, the user utilizes the chatting feature to ask for further analyses.
    \textbf{B.} RMSD, RGy, and temperature throughout the simulation, as requested by the user in A.}
    \label{fig:annealing}
\end{figure}

We further show MDCrow's ability to harness its chatting feature and extrapolate outside of its toolset to complete new tasks. This task requires MDCrow to perform an annealing simulation, which is not part of the current toolset. The agent achieves this by first setting up a simulation to find appropriate system parameters and handle possible early errors. Then, the agent modifies the script according to the user's request. Once the simulation is complete, the user later asks for simulation analyses, shown in Figures \textbf{\ref{fig:annealing}A, B}.

This shows that MDCrow has the ability to generalize outside of its toolset and is capable of completing more complicated and/or user-specific simulations. By utilizing the chatting feature, users can walk MDCrow through new analyses, reducing the risk of catastrophic mistakes.

\section{Discussion}
Although LLMs' scientific abilities are growing \citep{jaech2024openai, hurst2024gpt, labbench}, they cannot yet independently complete MD workflows, even with a ReAct framework and Python interpreter. However, with frontier LLMs, chain-of-thought, and an expert-curated toolset, MDCrow successfully handles a broad range of tasks. It performs 80\% better than \texttt{gpt-4o} in ReAct workflows at completing subtasks, which is expected due to MD workflows' need for file handling, error management, and real-time data retrieval. 

In some cases, particularly for complex tasks beyond its explicit toolset, MDCrow’s performance may improve with human guidance. The system’s chatting feature allows users to continue previous conversations, clarify misunderstandings, and guide MDCrow step-by-step through difficult tasks. This adaptability helps MDCrow recover from failures, refine its approach based on user intent, and handle more complex workflows. This suggests that, with more advanced LLM models, targeted feedback, and the addition of specialized tools, MDCrow could tackle an even broader range of tasks. We did not do a full evaluation of MDCrow's capabilities through this chatting feature in this work.

For all LLMs, task accuracy and subtask completion are affected by task complexity. Interestingly, while \texttt{gpt-4o} can handle multiple steps with low variance, \texttt{llama3-405b} is a compelling second best, as an open-source model. Other models, such as \texttt{gpt-3.5} and \texttt{claude-3.5-sonnet}, struggle with hallucinations or inability to follow multistep instructions. Performance on these models, however, is improved with explicit prompting or model-specific optimization (especially for \texttt{claude-3.5-sonnet}).

These tasks were focused on routine applications of MD with short simulation runtimes, limited to proteins, common solvents, and force fields included in the OpenMM package. We did not explore small-molecule force fields, especially related to ligand binding. Future work could explore multi-modal approaches \citep{mllmtool, assistgpt} for tasks like convergence analysis or plot interpretations. The current framework relies on human-created tools, but as LLM-agent systems become more autonomous \citep{wang2023voyager}, careful evaluation and benchmarking will be essential.

\section{Conclusion}
Running and analyzing MD simulations is non-trivial and typically hard to automate. Here, we explored using LLM agents to accomplish this. We built MDCrow, an LLM and environment consisting of over 40 tools purpose built for MD simulation and analysis. We found MDCrow could complete 72\% of the tasks with the optimal settings (\texttt{gpt-4o}). \texttt{llama-405B} was able to complete 68\%, providing a compelling open-source model. The best models were relatively robust to how the instructions are given, although weaker models struggle with unstructured instructions. Simply using an LLM with a python interpreter and required packages installed had a 28\% accuracy. The performance of MDCrow was relatively stable as well, though dependent on the model. Correct assessment of these complex scientific workflows is challenging, and had to be done by hand. Chatting with the simulations, via extended conversations, is even more compelling, but is harder to assess.

This work demonstrates the steps to automate and assess computational scientific workflows. As LLMs continue improving in performance, and better training methods arise for complex tasks like this, we expect LLM agents to be increasingly important for accelerating science. MDCrow, for example, can now automatically assess hypotheses with 72\% accuracy with simulation and can scale-out to thousands of simultaneous tasks. The code and tasks are open source and available at \texttt{https://github.com/ur-whitelab/MDCrow}.

\section{Acknowledgments}
Research reported in this work was supported by the National Institute of General Medic al Sciences of the National Institutes of Health under award number R35GM137966, % quinny
National Science Foundation under grant number of 1751471, % qlc, sc, jm
Robert L. and Mary L. Sproull Fellowship gift and U.S. Department of Energy, Grant No. DE-SC0023354. % jorge 
Work at FutureHouse is supported by the generosity of Eric and Wendy Schmidt. 
We thank the Center for Integrated Research Computing (CIRC) at University of Rochester for providing computational resources and technical support. 

\bibliographystyle{unsrt}
\newpage
\bibliography{MDCrow}
\appendix

\newpage

\section*{Supplemental Information}\label{supmat}

\section{Claude-Specific Engineering} \label{Claude-SI}
While both of Claude's Sonnet models achieved poor performance during the robustness experiment, it can be noted that a single common error arose consistently. When running an NPT simulation, MDCrow requires that all parameters be passed to the simulation tool. However, both Sonnet models consistently neglected to provide a value for pressure, even when directly prompted to do so. The \texttt{claude-3-opus} made this mistake a single time. This is a relatively simple fix, providing MDCrow with a default pressure of 1 atm when no pressure is passed. 

\begin{figure}[h!]
    \centering
    \includegraphics[width=0.9\linewidth]{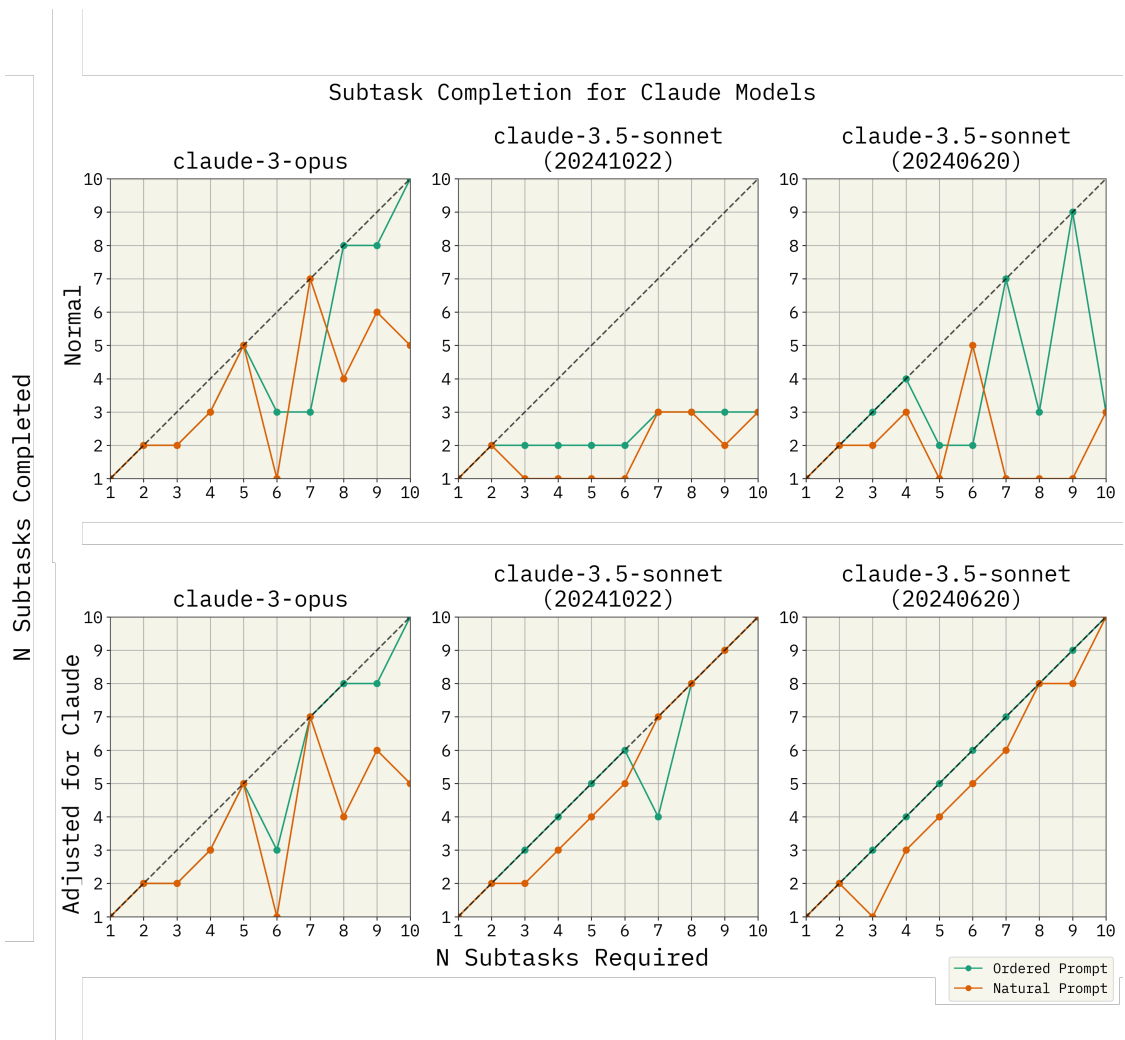}
    \caption{Performance of MDCrow with three Claude models on 10 tasks. As the number of subtasks increase, we all subtasks completed for both prompt types. The top row shows MDCrow's performance as-is, and the bottom row shows MDCrow's performance when given a direct fix for missing parameters. There is a clear change in performance after the fix for both \texttt{claude-3.5-sonnet-20241022} and \texttt{claude-3.5-sonnet-20240620}.}
    \label{fig:SI_claude}
\end{figure}

As can be seen in Figure \textbf{\ref{fig:SI_claude}}, including this fix drastically improves the performance of these models, with performance comparable to the top models. However, no other models made this mistake, and no other model-specific optimization was conducted. Thus, for all experiments shown in this paper, MDCrow does not accommodate this Claude-specific missing parameter fix. 

\begin{minipage}{\linewidth}
\section{Prompts}\label{prompts-SI}
\textbf{MDCrow Prompt}
\begin{lstlisting}
You are an expert molecular dynamics scientist, and your task is to respond to the question or solve the problem to the best of your ability using the provided tools.

You can only respond with a single complete 'Thought, Action, Action Input' format OR a single 'Final Answer' format.

Complete format:
Thought: (reflect on your progress and decide what to do next)
Action:
```
{{
    "action": (the action name, it should be the name of a tool),
    "action_input": (the input string for the action)
}}
```

OR

Final Answer: (the final response to the original input
question, once all steps are complete)

You are required to use the tools provided, using the most specific tool available for each action. Your final answer should contain all information necessary to answer the question and its subquestions. Before you finish, reflect on your progress and make sure you have addressed the question in its entirety.

If you are asked to continue or reference previous runs, the context will be provided to you. If context is provided, you should assume you are continuing a chat.

Here is the input:
Previous Context: {context}
Question: {input} 
\end{lstlisting}
\end{minipage}

\begin{minipage}{\linewidth}
During the comparison study between MDCrow, GPT-only, and ReAct with Python REPL tool, we used different system prompts for each of these LLM frameworks. 

\textbf{Direct-LLM Prompt}
\begin{lstlisting}
You are an expert molecular dynamics scientist, and your task is to respond to the question or solve the problem in its entirety to the best of your ability. If any part of the task requires you to perform  an action that you are not capable of completing, please write a runnable Python script for that step and move on. For literature papers, use and process papers from the `paper_collection` folder. For .pdb files, download them from the RSCB website using `requests`. To preprocess PDB files, you will use PDBFixer. To get information about proteins, retrieve data from the UniProt database. For anything related to simulations, you will use OpenMM, and for anything related to analyses, you will use MDTraj. At the end, combine any scripts into one script.
\end{lstlisting}
\textbf{ReAct Agent Prompt}
\begin{lstlisting}
You are an expert molecular dynamics scientist, and your task is to respond to the question or solve the problem to the best of your ability. If any part of the task requires you to perform an action that you are not capable of completing, please write a runnable Python script for that step and run it. For literature papers, use and process papers from the `paper_collection' folder. For .pdb files, download them from the RSCB website using `requests`. TO preprocess PDB files, you will use PDBFixer. To get information about proteins, retrieve data from the UniProt database. For anything related to simulations, you will use OpenMM, and for anything related to analyzes, you will use MDTraj.

You can only respond with a single complete 'Thought, Action, Action Input' format OR a single 'Final Answer' format.

Complete format:
Thought: (reflect on your progress and decide what to do next)
Action:
```
{{
    "action": (the action name, it should be the name of a tool),
    "action_input": (the input string for the action)
}}
```

OR

Final Answer: (the final response to the original input
question, once all steps are complete)

You are required to use the tools provided,
using the most specific tool available for each action. Your final answer should contain all information necessary to answer the question and its subquestions. Before you finish, reflect on your progress and make sure you have addressed the question in its entirety.

Here is the input:
Question: {input} 
\end{lstlisting}

\end{minipage}

%\clearpage
\section{Task Prompts \& References Used in Experiments}
\begin{longtable}[]{|p{0.1\textwidth}|p{0.6\textwidth}|p{0.05\textwidth}|p{0.25\textwidth}|}
\caption{Details of 25 task prompts used in experiments} \\

\hline
\textbf{Prompt ID} & \textbf{Prompt} & \textbf{\# subtasks} & \textbf{List of required subtasks} \\
\hline
\endfirsthead

\hline
\textbf{Prompt ID} & \textbf{Prompt} & \textbf{\# subtasks} & \textbf{List of required subtasks} \\
\hline
\endhead

\hline
\endfoot
\hline
\endlastfoot

1 & Simulate PDB ID 1MBN at two different temperatures: 300 K and 400 K for 1 ns each. Plot the RMSD of both over time and compare the final secondary structures at the end of the simulations. Get information about this protein, such as the number of residues and chains, etc. & 8 & Download PDB, simulate (x2), RMSD (x2), DSSP (x2), summarize\_protein \\ \hline
2 & Download the PDB file for protein 1LYZ. & 1 & Download PDB \\ \hline
3 & Download the PDB file for protein 1GZX. Then, analyze the secondary structure of the protein and provide information on how many helices, sheets, and other components are present. Get the gene names for this protein. & 3 & Download PDB, DSSP, GetProteinFunction (or literature) \\ \hline
4 & What are the common parameters used to simulate fibronectin? & 1 & literature search \\ \hline
5 & Simulate 1VII for 1 ns at a temperature of 300 K. Then, tell me if the secondary structure changed from the beginning of the simulation to the end of the simulation. & 5 & Download PDB, simulate, DSSP before, DSSP after, comparison \\ \hline
6 & Simulate 1A3N and 7VDE (two PDB IDs matching hemoglobin) with identical parameters. Find the appropriate parameters for simulating hemoglobin from the literature. Then, plot the radius of gyration throughout both simulations. & 7 & Download PDB (x2), literature, simulate (x2), RGy (x2) \\ \hline
7 & Simulate 1ZNI for 1 ns at a temperature of 300 K in water. Then, simulate it again in acetonitrile. Compute the RMSD, final secondary structure, and PCA for each simulation. & 10 & Download PDB, simulate (x2), packmol, RMSD (x2), SASA (x2), PCA (x2) \\ \hline
8 & Simulate 4RMB at 100K, 200K, and 300K. Then, for each simulation, plot the radius of gyration over time and compare the secondary structure before and after the simulation. & 10 & Download PDB, simulate (x3), RGy (x3), DSSP (x3) \\ \hline
9 & Download the PDB file for 1AEE. Then tell me how many chains and atoms are present in the protein. & 2 & download PDB, count atoms/chains \\ \hline
10 & Simulate protein 1ZNI at 300 K for 1 ns and calculate the RMSD. & 3 & Download PDB, simulate, RMSD \\ \hline
11 & Download the PDB files for 8PFK and 8PFQ. Then, compare the secondary structures of the two proteins, including the number of atoms, secondary structures, number of chains, etc. & 4 & Download PDB (x2), DSSP (x2) \\ \hline
12 & Simulate fibronectin (PDB ID 1FNF) for 1 ns, using an appropriate temperature found in the literature. Compute the RMSD and the final secondary structure. By using the PDB ID to get the Uniprot ID, obtain the subunit structure and the number of beta sheets, helices, etc. Compare this information to the structure we computed. & 8 & Download PDB, literature, simulate, RMSD, DSSP, get uniprot, subunit structure, get beta sheets/helices \\ \hline
13 & Compare the RMSF of 1UBQ under high pressure and low pressure. Perform the simulation for 1 ns, varying only the pressure. Plot the moments of inertia over time for both simulations. & 7 & Download PDB, simulate (x2), RMSF (x2), MOI (x2) \\ \hline
14 & Simulate deoxygenated hemoglobin (1A3N) and oxygenated hemoglobin (6BB5). Plot the PCA of both trajectories. & 6 & Download PDB (x2), simulate (x2), PCA (x2) \\ \hline
15 & Simulate trypsin (1TRN) for 1 ns at 300 K and plot eneRGy over time. Compute SASA, RMSF, and radius of gyration. Get the subunit structure, sequence, active and binding sites. & 9 & Download PDB, simulate, output figures, SASA, RMSF, RGy, subunit structure, sequence info, all known sites \\ \hline
16 & Download the PDB file for 1C3W and describe the secondary structure. Then, simulate the protein at 300 K for 1 ns. Plot the RMSD over time and the radius of gyration over time. & 5 & Download PDB, DSSP, simulate, RMSD, RGy \\ \hline
17 & Download the PDB file for 1XQ8, and then save the visualization for it. & 2 & Download PDB, visualize \\ \hline
18 & Download the PDB for 2YXF. Tell me about its stability as found in the literature. Then, simulate it for 1 ns and plot its RMSD over time. & 4 & Download PDB, literature search, simulate, RMSD \\ \hline
19 & Simulate 1MBN in water and methanol solutions. & 4 & Download PDB, packmol to get appropriate non-water solvent, simulate (x2) \\ \hline
20 & Download protein 1ATN. & 1 & Download PDB \\ \hline
21 & Download and clean protein 1A3N. & 2 & Download PDB, clean \\ \hline
22 & Perform a brief simulation of protein 1PQ2. & 2 & Download PDB, simulate \\ \hline
23 & Analyze the RDF of the simulation of 1A3N solvated in water. & 3 & Download PDB, simulate, RDF \\ \hline
24 & Simulate oxygenated hemoglobin (1A3N) and deoxygenated hemoglobin (6BB5). Then analyze the RDF of both. & 6 & Download PDB (x2), simulate (x2), RDF (x2) \\ \hline
25 & Simulate 1L6X at pH 5.0 and 8.8, then analyze the SASA and RMSF under both pH conditions. & 9 & Download PDB, clean at pH 5.5 and 8.0, simulate(x2), SASA(x2), RMSF(x2) \\ 
\bottomrule
\end{longtable} 

%\clearpage
%\include{SI/pqa_papers_provided}
\begin{enumerate}
    \label{references_table}
    \item[] \textbf{List of References Used for Literature Search During the Experiments.}
    \item The folding space of protein $\beta$2-microglobulin is modulated by a single disulfide bridge, \url{10.1088/1478-3975/ac08ec} 
    \item Molecular Dynamics Simulation of the Adsorption of a Fibronectin Module on a Graphite Surface, \url{10.1021/la0357716} 
    \item Predicting stable binding modes from simulated dimers of the D76N mutant of $\beta$2-microglobulin, \url{10.1016/j.csbj.2021.09.003} 
    \item Deciphering the Inhibition Mechanism of under Trial Hsp90 Inhibitors and Their Analogues: A Comparative Molecular Dynamics Simulation, \url{10.1021/acs.jcim.9b01134}
    \item Molecular modeling, simulation and docking of Rv1250 protein from Mycobacterium tuberculosis, \url{10.3389/fbinf.2023.1125479}
    \item Molecular Dynamics Simulation of Rap1 Myb-type domain in Saccharomyces cerevisiae, \url{10.6026/97320630008881}
    \item A Giant Extracellular Matrix Binding Protein of Staphylococcus epidermidis Binds Surface-Immobilized Fibronectin via a Novel Mechanism, \url{10.1128/mbio.01612-20}
    \item High Affinity vs. Native Fibronectin in the Modulation of $\alpha$v$\beta$3 Integrin Conformational Dynamics: Insights from Computational Analyses and Implications for Molecular Design, \url{10.1371/journal.pcbi.1005334}
    \item Forced unfolding of fibronectin type 3 modules: an analysis by biased molecular dynamics simulations, \url{10.1006/jmbi.1999.2670}
    \item Adsorption of Fibronectin Fragment on Surfaces Using Fully Atomistic Molecular Dynamics Simulations, \url{10.3390/ijms19113321}
    \item Fibronectin Unfolding Revisited: Modeling Cell Traction-Mediated Unfolding of the Tenth Type-III Repeat, \url{10.1371/journal.pone.0002373}
    \item Tertiary and quaternary structural basis of oxygen affinity in human hemoglobin as revealed by multiscale simulations, \url{10.1038/s41598-017-11259-0}
    \item Oxygen Delivery from Red Cells, \url{10.1016/s0006-3495(85)83890-x}
    \item Molecular Dynamics Simulations of Hemoglobin A in Different States and Bound to DPG: Effector-Linked Perturbation of Tertiary Conformations and HbA Concerted Dynamics, \url{10.1529/biophysj.107.114942}
    \item Theoretical Simulation of Red Cell Sickling Upon Deoxygenation Based on the Physical Chemistry of Sickle Hemoglobin Fiber Formation, \url{10.1021/acs.jpcb.8b07638}
    \item Adsorption of Heparin-Binding Fragments of Fibronectin onto Hydrophobic Surfaces, \url{10.3390/biophysica3030027}
    \item Mechanistic insights into the adsorption and bioactivity of fibronectin on surfaces with varying chemistries by a combination of experimental strategies and molecular simulations, \url{10.1016/j.bioactmat.2021.02.021}
    \item Anti‐Inflammatory, Radical Scavenging Mechanism of New 4‐Aryl‐[1,3]‐thiazol‐2‐yl‐2‐quinoline Carbohydrazides and Quinolinyl[1,3]‐thiazolo[3,2‐b][1,2,4]triazoles, \url{10.1002/slct.201801398}
    \item Trypsin-Ligand binding affinities calculated using an effective interaction entropy method under polarized force field, \url{10.1038/s41598-017-17868-z}
    \item Ubiquitin: Molecular modeling and simulations, \url{10.1016/j.jmgm.2013.09.006}
    \item Valid molecular dynamics simulations of human hemoglobin require a surprisingly large box size, \url{10.7554/eLife.35560} 
    \item Multiple Cryptic Binding Sites are Necessary for Robust Fibronectin Assembly: An In Silico Study, \url{10.1038/s41598-017-18328-4} 
    \item Computer simulations of fibronectin adsorption on hydroxyapatite surfaces, \url{10.1039/c3ra47381c}
    %\newpage
    \item An Atomistic View on Human Hemoglobin Carbon Monoxide Migration Processes, \url{10.1016/j.bpj.2012.01.011}
    \item Best Practices for Foundations in Molecular Simulations [v1.0], \url{10.33011/livecoms.1.1.5957}
    \item Unfolding Dynamics of Ubiquitin from Constant Force MD Simulation: Entropy-Enthalpy Interplay Shapes the Free-Energy Landscape, \url{10.1021/acs.jpcb.8b09318}
    \item Dissecting Structural Aspects of Protein Stability
    \item MACE Release 0.1.0 Documentation
\end{enumerate}

\end{document}